\def\year{2022}\relax
\documentclass[letterpaper]{article} 
\usepackage{aaai22}  
\usepackage{times}  
\usepackage{helvet} 
\usepackage{courier}  
\usepackage[hyphens]{url}  
\usepackage{graphicx} 
\def\UrlFont{\rm}  
\usepackage{natbib}  
\usepackage{caption} 
\usepackage{bm}

\usepackage{listings}
\usepackage[normalem]{ulem}
\usepackage{algorithm2e} 
\usepackage{xcolor}
\definecolor{dkgreen}{rgb}{0,0.6,0}
\definecolor{gray}{rgb}{0.5,0.5,0.5}
\definecolor{mauve}{rgb}{0.58,0,0.82}
\usepackage{tikz}
\usetikzlibrary{automata,positioning,arrows.meta}
\usepackage[caption=false,font=normalsize,labelfont=sf,textfont=sf]{subfig}
\usepackage{amsmath}
\usepackage{tikz}
\usetikzlibrary{automata,positioning,arrows.meta}

\title{
Overcoming Digital Gravity when using AI in Public Health Decisions\thanks{This manuscript is based on work funded by the Bill \& Melinda Gates Foundation, investment ID 52720.

IBM, the IBM logo, and other IBM trademark listed on the IBM Trademarks List are trademarks or registered trademarks of IBM Corp., in the U.S.and/or other countries.}
}
\author{
    Sekou L Remy,
    Aisha Walcott-Bryant,
    Nelson K Bore,
    Charles M Wachira,
    Julian Kuenhert
    \\
}
\affiliations{
    IBM Research Africa\\
    sekou@ke.ibm.com, awalcott@ke.ibm.com, nelsonbo@ke.ibm.com,\\ charles.wachira1@ibm.com, julian.kuehnert@ibm.com
}

\begin{document}

\maketitle

\begin{abstract}
In popular usage, Data Gravity refers to the ability of a body of data to attract applications, services and other data. 
In this work we introduce a broader concept, ``Digital Gravity'' which includes not just data, but other elements of the AI/ML workflow. 
This concept is born out of our recent experiences in developing and deploying an AI-based decision support platform intended for use in a public health context.  
In addition to data, examples of additional considerations are compute (infrastructure and software), DevSecOps (personnel and practices), algorithms/programs, control planes, middleware (considered separately from programs), and even companies/service providers. 
We discuss the impact of Digital Gravity on the pathway to adoption and suggest preliminary approaches to conceptualize and mitigate the friction caused by it.
\end{abstract}

\noindent Data Gravity is a term introduced in 2010 and refers to the observation that certain datasets tend to demonstrate an attractive pull on other resources around it \cite{mccrory2010data}.
This metaphor is based on the observation that as data accumulates, there comes a point at which it becomes difficult to move.
From that point it is easier to move services and applications towards the data than moving the data to the services or applications.
Moreover, other (smaller) data sources are also easier to move, so they too are pulled into the so-called orbit of the data in question.

In this work we introduce a broader term, \textit{Digital Gravity}, in which we consider not just data, but other elements of the AI/ML workflow.
We seek to consider the entire set: data, compute (infrastructure and software), DevSecOps (personnel and practices), algorithms/programs, control planes, middleware (considered separately from programs), licensing choices (for both code and data), and even companies/service providers.
Extending the original metaphor, instances of each of these concepts have some mass, and depending on the context or usecase can generate gravity large enough to draw in instances of the other concepts.
An example of this might be that because of the policies within an Agency, all computation must be performed within a specific compute infrastructure.
Accordingly, certain operating systems must be used; data must be encapsulated within a particular data paradigm; models must be implemented and managed using a particular middleware; and compute must be deployed using a particular platform.
This is just one possible chain of implications of the base policy directive of the Agency.

To discuss this more plainly, we will present an implementation of decision support with AI which was developed to support users in a public health context.  This implementation was deployed in a limited proof of concept in a research context and we consider what was learned during its implementation and the ways that it was influenced by Digital Gravity in real world settings. 
Finally, we consider what steps could be taken to overcome or compensate for the challenges therein.  Since our specific approach is an exemplar from a broader class of applications, we will also demonstrate how the concepts apply more broadly.

\section{Background}
To ground the concept of Digital Gravity, we present a concrete example of an approach to inform decision support processes in a public health context.
Our work is complementary to the framing espoused in \cite{schuck2021information}, in that we consider the impact of gravity on the configuration of a system, but the presence of adversarial elements would be addressed through the use of frameworks or platforms to enable the deployment of critical components with the overall structure.
Our use of physical metaphors is adjacent to the way those authors consider [digital] information maneuverability.

\subsubsection{Problem}
Public health professionals are expected to engage in evidence-informed decision making to advise their practice \citep{petticrew2004evidence, yost2014tools}.
As indicated in \citet{nutbeam2001evidence}, evidence is required from a variety of sources including expert knowledge, existing domestic and international research, stakeholder consultation, and even assessment of existing policies.
While it is rational to expect evidence to be used, challenges in evidence based decision-making are ever present. 
Today, very few decisions have complete information, the simplest decisions often have innumerable outcomes which are ultimately uncertain. 
For high impact decision making, such as epidemic planning, though of great importance, it is not possible to evaluate all possible options, or completely characterize their uncertainty.
Many turn to modeling as a mechanism to generate insights about what could potentially happen, and consider model output as informative in the decision making process.
It is well known that all models can be considered to be `wrong' \citep{ioannidis2020forecasting}, and in some cases have led to ill conceived actions being considered, but it has also been shown that when properly contextualized, even these incorrect models can be useful \citep{holmdahl2020wrong}.

\subsubsection{Ecosystem}
There are very few settings where high impact decisions are determined by a small group of individuals in the same organization, or at a single point in time.
Instead, as we have seen during the current global pandemic there are multiple stakeholders and a plurality of skillsets required to even have a chance at a successful intervention.
Further, these resources are spread across multiple organizations and the composition needed changes over time.
Accordingly there is a decision-making ecosystem.

In some countries these ecosystems are completely self-contained with occasional external reporting to bodies like the World Health Organization.
More often however there is cross border interaction, and even in the in-country context communication and collaboration across sub-national boundaries can come with as many of the challenges as international engagements.
In our case we consider a multi-country ecosystem including governmental, university, private sector, and NGO partners.

\subsubsection{Approach}
We demonstrate an approach to harness epidemiological models and generate evidence which can inform decision making.
We utilize Machine learning algorithms from a range of classes (from RL, to Optimization, to Planning).
Our aim is not to focus on model validity as defined in \citet{petticrew2004evidence}, nor the non-technical issues related to adoption of the insights generated from these models, \citet{liverani2013political}, nor even to adjudicate which model (or algorithm) is `best'.
Our goal is to show the community how models can be connected with these ML algorithms in a flexible manner which will support integration of multiple classes and implementations of models.
What we champion here will also provide a mechanism to realize a repeatable infrastructure to generate insights, not just for COVID-19, but for any decision making process which can be informed by multiple models and data sources. 

\section{Methodology}
Learning which sequences of interventions may be best to consider is rooted in an approach introduced in \citet{notnets}.
Our team found the approach well suited to access models at scale with transparency and trust (see Figure \ref{fig:arch_new}).
Specifically, this framework enables API mediated access to a suite of models.
Containerized models are deployed on-demand, and configured at runtime to use the input parameters defined by the user.
The framework marshals the use of the model and the associated data in a manner that gives both the users and owners of these assets confidence in the output which is generated.
Moreover, the framework permits data collected by different algorithms, and possibly even at different times, to be shared. 
Every additional execution of an algorithm contributes to the pool of knowledge which exists about the models.
Finally, the framework also supports the flexible integration of input into a format which can be consumed by the specific models which are available.
Figure \ref{fig:dataflow} presents another view of the relationship between the elements within the platform by considering how data and requests flow when generating and consuming results.

\begin{figure}[t]
\centering
\includegraphics[width=.9\columnwidth]{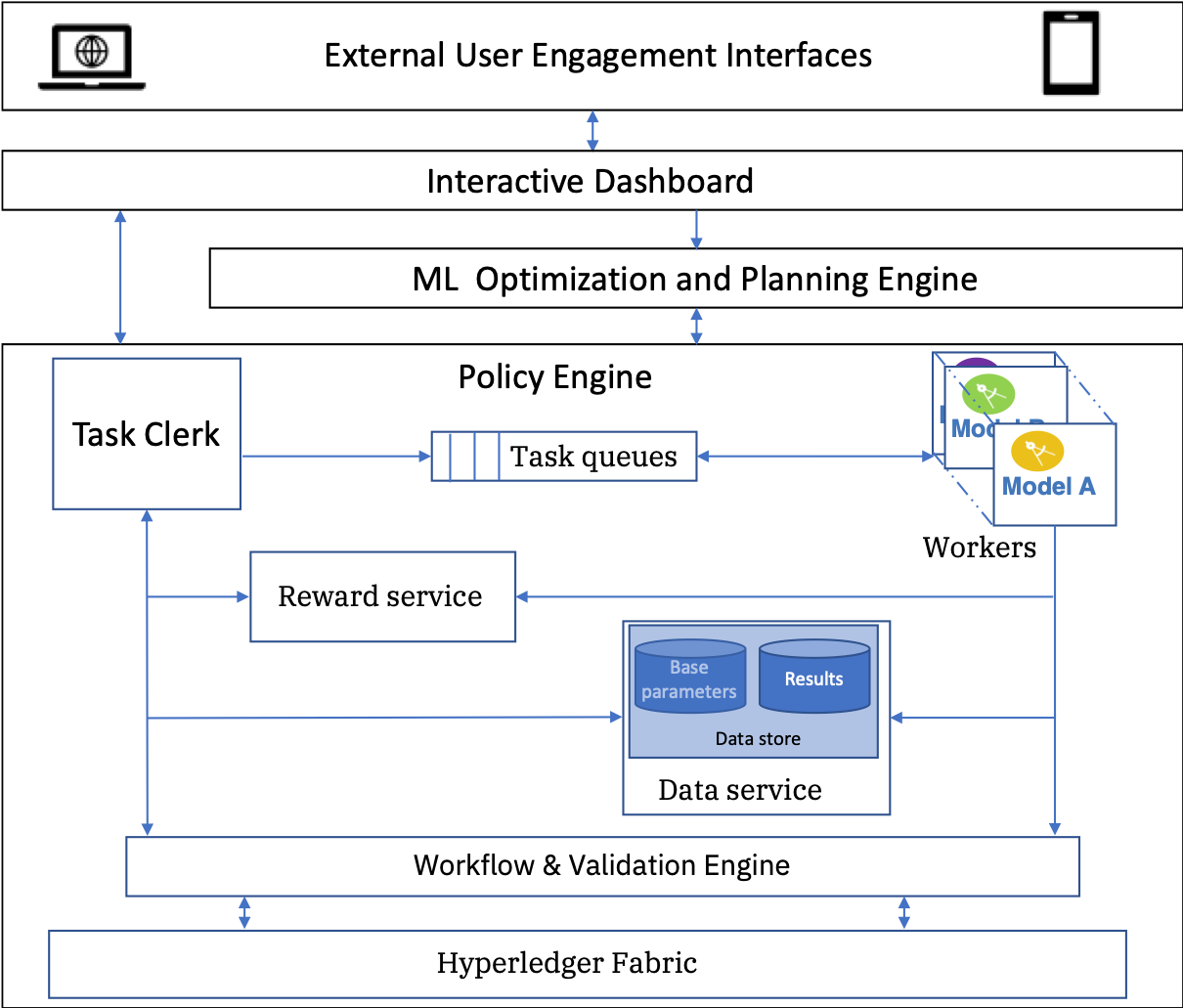}
\caption{Conceptual architecture of the platform used as the basis for this work.}
\label{fig:arch_new}
\end{figure}
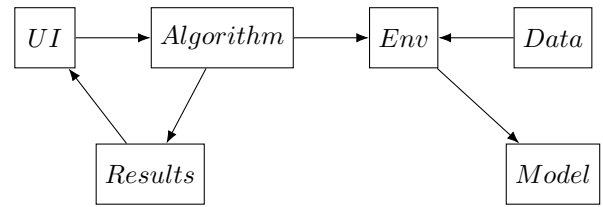
\begin{figure}[t]
\centering
\begin{tikzpicture}[node distance=1cm, auto,
    >=Latex, 
    every node/.append style={align=center},
    int/.style={draw, minimum size=.8cm}]

   \node [int] (dashboard)             {$UI$};
   \node [int, below=of dashboard, xshift=1.4cm] (results)             {$Results$};
   \node [int, right=of dashboard] (algo)             {$Algorithm$};
   \node [int, right=of algo] (env) {$Env$};
   \node [int, right=of env] (data) {$Data$};
   \node [int, below=of data] (model) {$Model$};
   \coordinate[right=of model] (out);
   \path[->] (algo) edge node {$$} (env)
             (data) edge node {$$} (env) 
             (dashboard) edge node {$$} (algo) 
             (results) edge node {$$} (dashboard) 
             (algo) edge node {$$} (results) 
             (env) edge node {$$} (model);
\end{tikzpicture}
\caption{Primary flows within the considered use case.
The key abstraction presented in \cite{notnets} wraps domain models in OpenAI Gym environments.
This separation between domain models, data, and algorithms enables more flexible interactions between these three types of resources.
}
\label{fig:dataflow}
\end{figure}

\subsubsection{Deriving a Transmission Rate}
In one treatment, a user may desire to learn what transmission rate might be inferred from the case data for a location of interest.
This information can be generated by tuning a trusted model to the available case data, for a specified timeframe.
When tuned properly, the transmission rate can be observed directly from the model.
To perform this task, the user needs access to a model which utilizes the transmission rate in the generation of time series output, and they need access to real-world data capturing the properties explicitly defined in the model's metadata.

Tuning was performed by coupling an algorithm, for example Bayesian Optimization, with an environment which maps parameter estimates to a scalar value ($g :R^n \mapsto R$).
This environment derives the scalar value from the output of the underlying model and the provided case data, but through encapsulation keeps the calculation details of independent of both the model and the algorithm.
For currently on-boarded COVID-19 environments, the scalar values used in calibration were all calculated using the Normalised Root Mean Square Error (NRMSE) which is defined in (\ref{eq:nrmse}). 
In this case $y$ is the model output sampled at time $t$ and $\hat{y}$ is the desired value of this output. 
We use $n$ samples of model output to calculate this measure and calibration will minimise this objective (See Algorithm \ref{bayes_opt}). 
\begin{equation}
g = \frac{\sqrt{\frac{1}{n} \sum_{t=1}^{n} ( y_t - \hat{y}_t ) ^2}}{\max_t(y_t) - \min_t (y_t)} 
\label{eq:nrmse}
\end{equation}
Bayesian Optimization builds $\hat{g}$, an iteratively more accurate surrogate of the objective $g$ around the places which could minimize it.
A multidimensional Gaussian Process is the key element which makes this possible.
This is the means through which the values which minimize $g$ (and thus calibrate the model) are learned.

For this work $\hat{g}$ is approximated using (\ref{eq:value}) with $\beta = 2$, effectively fixing a 95\% confidence interval on samples from the parameter space. 
Finally, $\mu$ and $\sigma$ are the mean and standard deviation for the model evaluated with particular set of parameters, $x$.
 
\begin{equation}
    \hat{g}(x) = \mu(\bm{x}) + \beta*\sigma(\bm{x})
    \label{eq:value}
\end{equation}

\begin{algorithm}
\caption{Bayesian Optimization, one algorithm implemented to enable calibration and learning intervention plans.}
\label{bayes_opt}
\KwResult{$\hat{x} = \operatorname*{argmin}_{\bm{x}} \hat{g}(\bm{x})$}
 Initialise: 3 random Parameters $\bm{x}_{1,2,3}$;\\
 \quad Surrogate  parameters $\mu_0 = 0$, $\sigma_0,l_0$;\\
 \For{i = 4,5,6 ...}{
 $\bm{x_{i}} = \operatorname*{argmin}_{\bm{x}}\hat{g}(\bm{x})$; \\
 $R_i$ = g($\bm{x_{i}}$) \\
 Return: $R_i$ \\
 Update Posterior; mean $\mu_{i}(\bm{x})$; variance $\sigma_{i}(\bm{x})$; and length scale $l_i$
 }
\end{algorithm}

Using the UI (Interactive Dashboard element) a user expresses their intent to perform this task and which model and data they prefer.
This intent is received by an algorithm which identifies which environments are available for consideration.
The environment is an abstraction which permits flexible and principled coupling between the available algorithms and models.
The algorithm then sends the selected environment(s) payloads which define what information is ultimately sent to the model.
Based on the request from the user, the algorithm is also responsible for specifying requests for the data needed to perform these model runs which the environment consumes and executes.
The algorithm generates numerous (for many cases on the order of thousands) of calls to the environment, each of which is associated with a model run.
For each model run, the result is provided back to the environment which performs some processing and the processed information is returned to the algorithm.
For this case of tuning the model, the algorithm's results are the best sets of parameters.
This information is then relayed, stored and made available to the UI for the user to consume.

\subsubsection{Transmission Rate in a Different Location}
If the user seeks to perform this transmission rate analysis for another location, using the UI they can check if there is data already within the platform and that they have access to it.
At present basic case data is available for most countries in the world, and administrative-level-1 (e.g., states in the United States of America) for a small number of countries.
If they would prefer other data which are available for an existing country or for another geography (e.g, another country or another administrative-level) it can be shared with our team and subsequently on-boarded to the platform once the appropriate data agreements have been processed.
The on-boarding process involves verification of metadata and units of the data.
We do not modify the source data, we add it to the catalog which is available for use within the platform.
Once the required data are available, the user simply chooses it, identifies the date range of interest, and the rest of the calibration process remains the same as previously described.

\subsubsection{Using a Different Model or Algorithm}
If the user seeks to perform this analysis but instead use a different model of COVID-19 or a different algorithm to tune the model, they can check within the UI dropdown values for the descriptor of alternatives (and whether they have access to them).
We currently have a few types of each of these available for selection.
For each model, in earnest what the user is selecting is truly an environment and there can be a many to one mapping between a model and environments.
The plurality of the environments permit these users to engage with the underlying models within the bounds specified by their creators.
This permits the model creator to set controls on the available parameter sets and their respective ranges.

As with data, if there are other model or algorithm sources, these can be on-boarded to our platform and the assets would be containerized an incorporated into the fold once the appropriate software agreements have be processed.
Once the required algorithms or models are available, the user simply chooses them, specifies the data of interest, and the rest of the calibration process remains the same as previously described.

\subsubsection{Generating Intervention Programs}

If instead of learning the $transmission\_rate$, the user instead wanted to generate recommendations for sequences of interventions for the same location based on calibrated models, the user would perform a very similar flow.
In this case, the environment would be different since the objective function is different and this is one of the elements encapsulated within the environment \cite{pmlr-v123-remy20a}.
The remainder of the elements in the flow remain the same.
Examples of environments used in this work have been released via a public GitHub source repository\footnote{https://github.com/IBM/ushiriki-policy-engine-library} which serves as both an installation point and templates for continued extension.

\begin{figure*}[t]
\centering
\subfloat[font=footnotesize][Agent Based Model]{
\includegraphics[width=.32\textwidth]{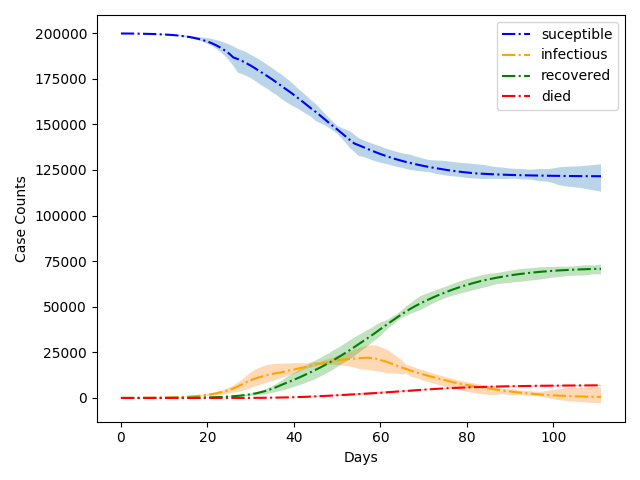}
\label{fig:covasimcalib}
}\hfill
\subfloat[font=footnotesize][Compartmental Model with Stringency]{
\includegraphics[width=.32\textwidth]{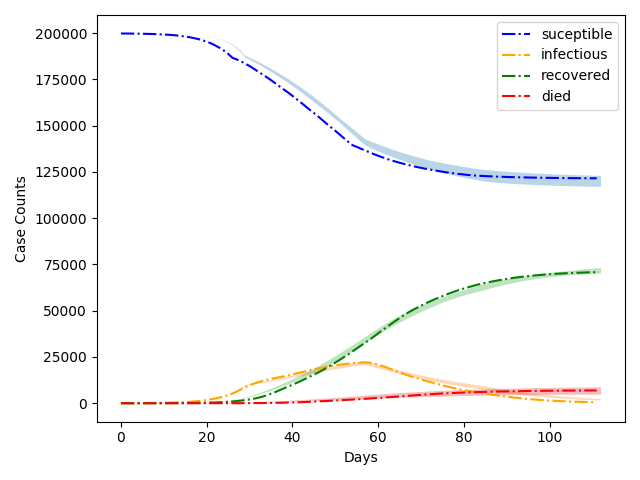}
\label{fig:sirdcalib}
}\hfill
\subfloat[font=footnotesize][Compartmental Model]{
\includegraphics[width=.32\textwidth]{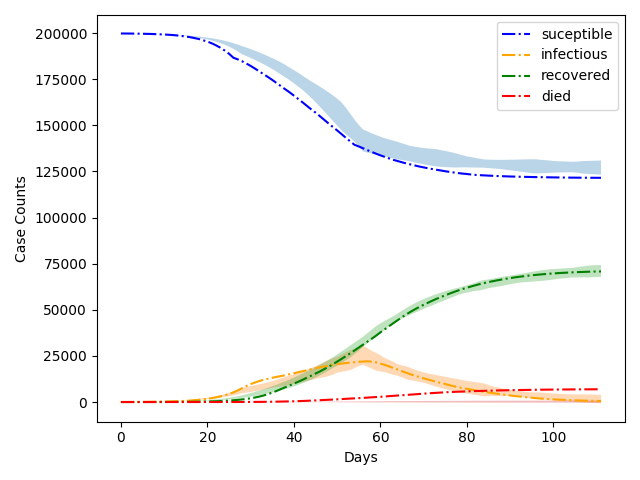}
\label{fig:sirdsicalib}
}
\caption{Calibration of three environments to the same case data (in this case using more data than is typically available).
In each calibration, the results consider learning 20 sets of parameters.
}
\label{fig:calibrations}
\end{figure*}

\subsubsection{More Broadly}

For the larger project in which this work is situated, we select models provided by modelers from different communities. 
We have also successfully deployed and utilized algorithms from multiple domains of the Machine Learning spectrum. 
From Reinforcement Learning \citep{watkins1992q, keerthi1994tutorial, mnih2013playing} to Optimization  \citep{Srinivas2009, Contal2013}, they have all been implemented utilizing the \textit{step} method of the environment to access the objective function (and in some cases the observations of the model states) when presented with an intervention to evaluate.
In this manuscript we will only highlight the use of two types of models:  compartmental models \citep{brauer2008compartmental, edlund2011comparing} and agent based models \citep{macal2009agent, Kerr2020.05.10.20097469} as they are both widely used by epidemiologists across a variety of domains.
For these instances of COVID-19 models referenced in this work, the countermeasures (or interventions) are captured as sequences of parameter values.
In the first framing, we permit the actions (e.g. physical distancing, mask wearing, vaccination) to directly change model parameters like transmission rate, recovery rate, and the death rate.
In an alternate framing, we also link model parameters to the Stringency Index \citep{hale2020variation}, a metric which captures the impact that a particular set of countermeasures has on disease progression. 
These framings represent distinct implementations of models either for model predictions or assessing the impact of control policies.

\section{Results}

\begin{figure}[t]
\centering
\subfloat[][Lenient]{
\includegraphics[width=.45\columnwidth, height = 3cm]{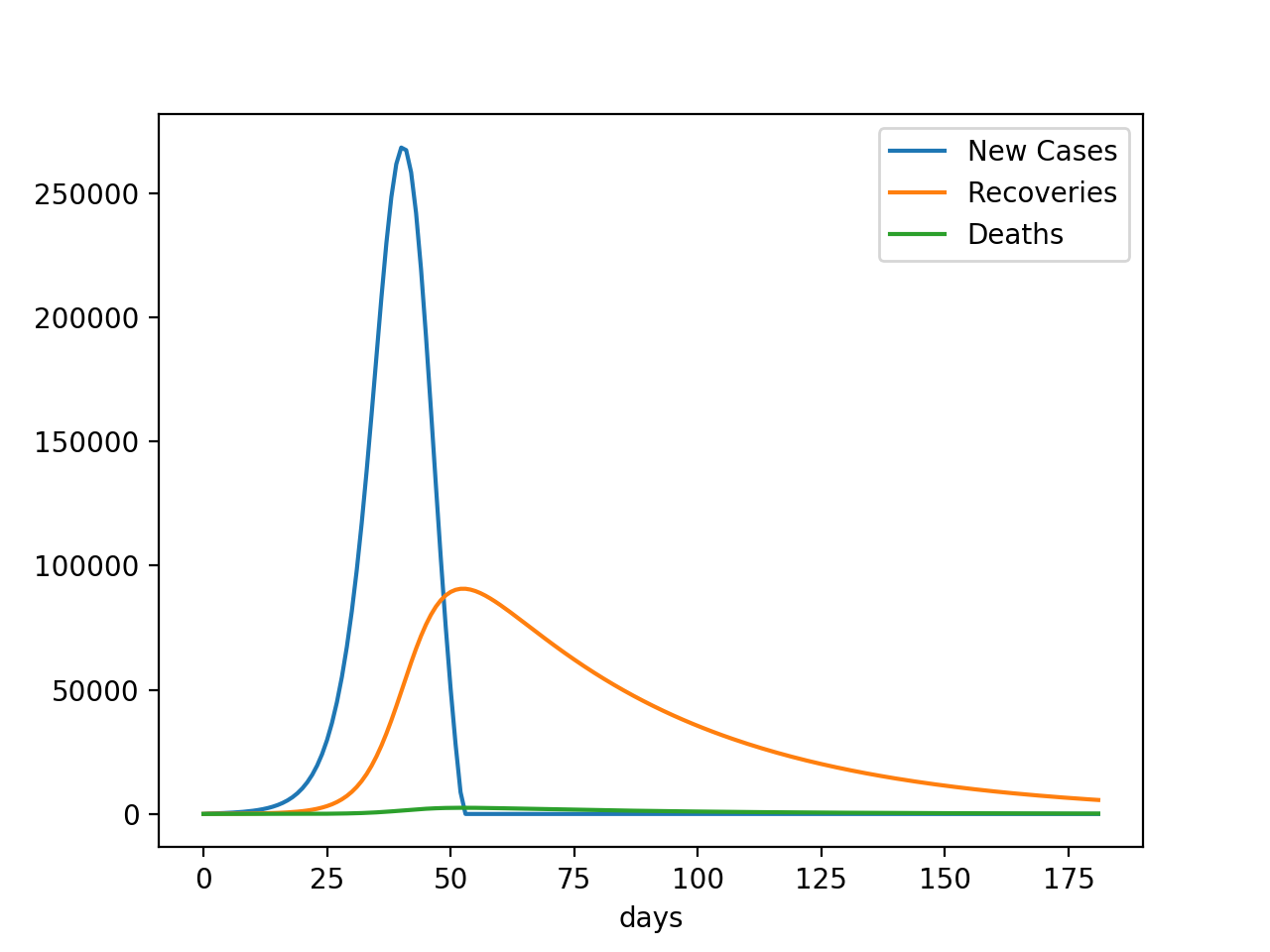}
\label{fig:90}
}\quad
\subfloat[][Stringent]{
\includegraphics[width=.45\columnwidth, height = 3cm]{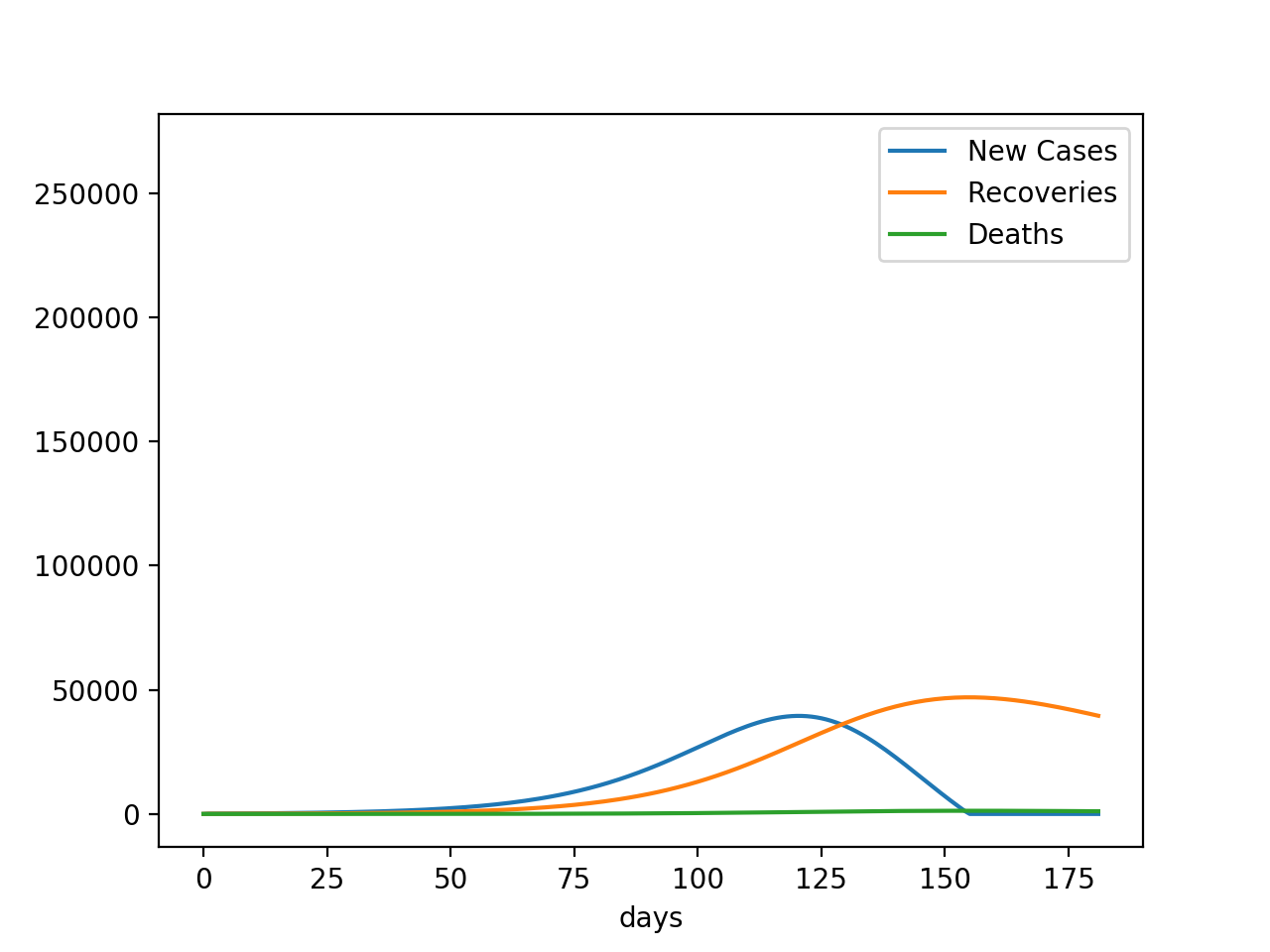}
\label{fig:09}
}
\caption{Two examples of deterministic policies of a single countermeasure for the simulated timeframe.
Could a reactive policy have performed ``better'' according to a chosen reward function?
}
\label{fig:policies}
\end{figure}

\textit{What are the most cost effective ways to ``flatten the curve''?
If there is a fixed amount of money which can be allocated to interventions over the next three months, what sequence would be best?
If you wish to simplify your policies, which reduced sets of interventions are most impactful?}

To address any of these questions using model derived evidence, one must first calibrate the model to data for the location in question, and then one can utilize an environment which provides the reward functions relevant to the particular question.
As indicated previously, the same algorithms can be applied with the multiple algorithm types and model types (through the use of the respective environments).
Moreover, for each model, there were environments for calibration and optimization with various objective functions.
A subset of the combinations is included herein for brevity.
Figure \ref{fig:calibrations} shows the results of calibrating the parameters of three different models to the same data.
In each case the algorithm was able to identify reasonable fits, even though the structures of the models were quite different.

In the case of calibration, the objective function is the error between the data and the observed model output.
The actions permit the algorithms to set the values of the calibration parameters.
By changing the objective function to represent different properties, e.g. the number of cases, and changing the actions to modify the transmission rate, another environment can be created permitting the same models to help consider the problem of flattening the curve.
In this case, the algorithm's role is to find the best set of interventions (the sequence of transmission rates) which minimize the number of individuals who contract the disease.

In this case, the environment's states are the values from each of the compartments (or the fraction of the population in a given state); 
the rewards produced are the cumulative incidence for the two week period which each step is assessed for; 
and the actions are integers in the range $[0,99]$. 
The overall goal is thus to minimize the total number of cases by learning when to vary the stringency of the interventions implemented.
In Figure \ref{fig:policies}, we show the state trajectories (incidence) for two sequences of actions.
The first was a very stringent disease management response (restrictive and invasive Non-Pharmaceutical Interventions) which was learned, while the other may be considered lenient.
In both cases there are individuals in the population who contract the disease, however under the stringent policy, there are much less who do.
Also, under this policy, the peak number of those who are infected at the same time is also lower.
This might be considered as an example of the impact of ``flattening the curve''.
Using this type of approach, and by changing the reward function, it would be possible to raise the question of whether there are other intervention programs (sequences of actions) which could have flattened the curve with a lower societal cost.

As alluded previously, by changing the objective function defined within the environment while keeping all the actions the same we can also consider the impact of sequences of actions under a different rubric.
In this case we could quantify the impact of those interventions on the community.
Executing interventions have a cost and can also be a burden on the population, but there's also a cost if individuals in the populations get infected.
Individuals can be infected even when the most stringent interventions are put into place.
Health economists have defined the notion of Years of effective Life Lost as one such measure to quantify the impact of an illness.
In environments used in this work we implement what they have outlined, and also include the cost of an intervention inferred from the newspaper sources.
Such costs are data which at present are embedded directly within the environment object, however could have been accessed from other location-specific data sources.

\section{Digital Gravity}
Our team developed and deployed the elements of this work and worked with an ecosystem considering the effects of government interventions on the spread of COVID-19.
We also used the same platform to consider the impact of AI on decision support with malaria domain models.
Both of these experiences enabled engagement with epidemiologists, bioinformaticians, public health practitioners, program managers and researchers at NGOs, scientists and administrators both from universities and government service, as well as researchers from machine learning and data science within the academic community.
It is from this context that we present our observations.

In Physics, gravity is a phenomenon where all things with mass (or energy) are attracted to each other.
The larger an object's mass, the greater the (gravitational) pull it has on the others.
In the original framing of Data Gravity, datasets with large mass were observed to demonstrate a pull on other resources.
As introduced in this work, Digital Gravity refers to the observation that many types of digital assets and processes also demonstrate this interaction with other resources around them.
In this work we have introduced the concept of Digital Gravity, but we also now present conceptual approaches to mitigate the impacts of the phenomenon on the adoption of AI.
After this, we will provide examples of both the effect and how measures were implemented during the course of our engagement as a result of its presence.

\subsection{Reducing the Negative Effects of Digital Gravity}
As gravity is a force, and forces have both magnitude and direction, conceptually these specify two dimensions through which gravity can be experienced.
Accordingly there will be some efforts which can address one or both of these dimensions of influence.


We identified three mechanisms to control the influence of the negative effects of Digital Gravity, and they are all related to inertia.
Inertia is the resistance to a change in motion, and for our context, this change in motion is generating a result of the act of using AI to generate a result.  

\subsubsection{Overcome Inertia}
In a physical sense, it's possible to implement a change in the system which makes it possible to overcome inertia.
It still will take some force to move the object (i.e., accomplish useful work); however if the changes were not made, it would not have been possible at all.
In the context of Digital Gravity, this could be realized through the use of a concept like either  container or OS virtualization which enables moving an application from execution in one physical location to running in a different one. 
This is a mechanism to move an algorithm or some processing function close to the data (i.e. the basic Data Gravity context).
It is also possible for the same principle to be applied when data is moved from one location to another through the use of concepts like Network File Systems or Object Stores.

It is possible, and increasingly the case that the effects of Gravity can impact multiple independent elements of a single system.
Data or compute may be housed in more than one physical location and may need to be moved against the force of their significant respective gravity.
This can be considered a case akin to the use of casters which makes it easier to move in more than just one direction.
There are a growing number of standards and psuedo-standards which exist within the Tech community, so in general this mechanism is largely addressed if developers make effective use of the appropriate standards in the field, although there is a clear need for more.

\subsubsection{Reducing Static Friction}
In the previous section we identified methods which make it possible to move digital resources as a means to overcome inertia and indicated a force was still applied once the system was modified, but the magnitude of that force might be quite large.
This second mechanism focuses on reducing the magnitude of that force.

In the context of an AI deployment, this could be seen through the use of a database which is widely available across multiple cloud providers, is very easy to run and manage on a local computer, and is taught broadly in academic institutions or is easy to learn/use.
This means that if for some reason there is need for the deployment of AI to utilize a specific deployment, it would be easier to do so, or in the case that support personnel need to be hired in locations that are off the beaten path, the pool of people familiar with that database is maximized.

Another way that this can be realized in the context of an AI deployment is through the use of some type of middleware which can manage operation of application in a multi-cloud or hybrid-cloud setting.
This is especially valuable in the case where some resources must necessarily be deployed in a heterogeneous context either due to legacy collaborations, vendor lock-in on a subset of the system components, or overall resiliency and robustness requirements.
A single interface makes it easier, and reduces the coefficient of friction experienced in managing the overall deployment.

Finally, when sharing or movement of resources is possible there are generally concerns about the governance and provenance of people, workflows, and resources.
In the context of deploying AI applications, the use of recognized and auditable tools to help manage these changes.
These tools provide the means for appropriate regulatory oversight and reduce the load on the DevSecOps personnel as they perform their job function.

\subsubsection{Re-purposing Kinetic Friction}
In physics there's a concept known as the Bernoulli Principle which helps sailboats use the force of the wind, and they're even able to do so when the wind is blowing opposite to the direction the boat needs to travel.
This is possible because sailboats also have some type of modification which opposes motion in the direction lateral to the boat's motion.
This combination of harnessing the force of the wind and high lateral inertia makes it possible to use the force to do useful work, even though it is acting in the ``wrong'' direction.

In the context of an AI deployment, when data gravity is high, it is sometimes possible to add more data, or to process that data in a manner that provides benefit.
These actions do not reduce the friction associated with the use of the data, but instead they can generate value which would not have existed without that data.
One example is that the data could be processed and new data generated and stored (i.e. increasing the mass), but in the case that there's a request for the processed data, there's a very low overhead to provide it to the user.
A variant of this example is using that processed data and generating a surrogate model.
This model becomes a stand-in for the data which unlike the initial case which merely provides quick look-ups, can perform inferences and provide responses for requests for data which do not exist within the corpus at all.
Just as in the case of the sailboat, work is able to be done by utilizing the force (in our case due to gravity not wind) and introducing lateral friction (providing lateral inertia).
If the gravity was small, the value which can be extracted by modifying the system would not be very meaningful, so even if it's acting in a contrary direction when that gravity is large it is possible to be redirected to become effective.

\subsection{Example Sources of Digital Gravity}
To make this concept concrete, we present a few examples we observed during our exercise of developing and deploying the presented system.
The sources of Digital Gravity in each example come from multiple sources, but we've revised them a bit for narrative clarity.

\subsubsection{The Technical Skillsets Within the Ecosystem}

Since the technologies implemented in this work were not part of the common user experience in a public health context, our team had to perform training as a form of sensitization prior to consideration.
This instruction included the science behind the algorithms implemented, briefings on the programming languages used, and introductions to the broader space of Cloud Computing, and AI/ML.
This content was not just for partners, but also their partners as well, and the process of providing this training was impacted by the technical exposure and computational infrastructure available with the ecosystem.

One approach needed to address this effect was to overcome inertia.
We deployed resources and provided synthetic data within a completely managed cloud infrastructure so that users would not have to install (m)any things locally, and also so that the installation and overall experience would be consistent, especially across borders.
We also chose resources which are also widely used in other cloud and non-cloud contexts, so that users could take what was learned and apply after our engagement.




\subsubsection{Our Team's Previous Experiences Analyzing Data}
Due to our team's recently published work generating a platform using NLP technology to analyze COVID-19 data at scale \cite{suryanarayanan2021ai}, other earlier work using Machine Learning to analyze longitudinal health data \cite{ogallo2020identifying}, and our engagement with a diverse community of researchers and students \cite{pmlr-v123-remy20a} some members of the ecosystem were motivated to consider extensions of the originally planned work.
This extension connected our team with researchers in neighbouring countries who were considering research questions which could be augmented by our experiences with relevant data, our technical capacity, our geography, and our experience performing technical work and meeting deliverables with geographically dispersed teams. 

In addition to the training mentioned in the previous section, since the effect now required different level of data within the platform, our team co-created a tool called the WNTRAC Curator\footnote{https://ibm.github.io/wntrac-curator/curator} to enable the new  collaborators to ensure the type of information they needed was available for their country within the platform.
The process resulted in derivative benefits for all countries, as the tool was made available as a public good complementing the original data platform.
Working together in this way not only were they able to more comprehensively generate results to address their research question, but our team was able to extend the reach of the developed assets, and also to integrate COVID-19 models with more contextually relevant data for our setting.
The resulting collaboration has led to additional opportunities for the new collaboration, and this provides a case where digital gravity has resulted in increased value through synergy.

\subsubsection{The Need to Build on Prior Investments}
In many cases, the partners insisted on working within the context of existing, trusted collaborations, and with familiar or well-known resources.
Beyond expected resistance to new things, these partners explicitly expressed that additional value is derived by re-using or amplifying the use of existing assets.
There's also a strong incumbent culture which resulted in using models and specific data instruments.

One approach needed to address or manage the impact of this effect was to overcome inertia.
Our platform was designed in a manner which enabled containerized deployment of code for models, and these models were executed in an environment very similar to the ones in which they were developed.
Data elements were accessed either explicitly via a data service, or via the API for the database where it was stored, in both cases the endpoints were either on publicly accessible host, or deployed in the same namespace as the model which needed access.
We also utilized permissioned and audited access to both models and data, all within setting of the same the deployed infrastructure, no matter the source of the data or the physical location where it was housed.

Our approach also permitted multiple independently developed models to co-exist within the same platform and to be accessed via the same abstracted interface.
Also, we supported data access via URIs which were housed in a plurality of contexts, and the environment(s) defined to bridge particular algorithms, models, data were implemented as required for the particular need.
This was one means of overcoming the static friction involved in considering a new model, or algorithm, or datasource.

We learned that approaches needed to address this effect, and specifically there was need to reduce the static friction even further.
There was a need for more granular controls on data and model use in both physical deployment location and virtualized deployments.


\section{Summary and Future work}
In this work, we introduce the concept of Digital Gravity and ground it in lessons learned during a recent research deployment of a decision support platform.
This AI powered platform utilized several elements of modern software engineering workflows, each of which both  induced and experienced forces due to gravity in varying directions and with varying magnitudes.
The forces which were observed during this deployment were not just as a result of Data, and using selected examples we show both how they were manifest, and what steps were taken to minimize the negative elements of their presence.
Taking a step back, we also shared more broad examples of ways which in general could reduce the the negative effects of digital gravity.

This work is squarely rooted in the public health context of decision support with infectious disease (both COVID-19 and brief discussion of Malaria), however the lessons learned apply to the wider context of systems which need to marshal the use of data, models, algorithms (general processing), and compute in a distributed setting with stakeholders partitioned across multiple organizations cooperating (or cooperating+competing) across boundaries.

There are many more lessons to learn, but it is already clear that there are critical gaps.
For one, in our implementation the strategy of using a centralized (within one provider), managed platform was not the best decision for key stakeholders and sponsor users.
This was not identified during the initial process of engaging users during the early phases of design, however there seems to be value further reduce static friction by implementing the ability to virtualize and more flexibly manage the storage and sharing of data and models ``across the edge.''
In earnest, there seems to be need to frame the pathway to adoption of AI in this context as one very similar to the case considered in next generation edge computing deployments.



\bibliography{refs.bib}
\end{document}